%% file: main.tex
\documentclass{article}

\usepackage{microtype}
\usepackage{graphicx}
\usepackage{subcaption}
\usepackage{booktabs} 
\usepackage[table]{xcolor}

\usepackage[accepted]{icml2026}

\usepackage{amsmath}
\usepackage{amssymb}
\usepackage{mathtools}
\usepackage{amsthm}

\input{configs}
\input{math_commands}

\usepackage{multirow}
\theoremstyle{plain}

\theoremstyle{definition}

\theoremstyle{remark}

\usepackage[textsize=tiny]{todonotes}

\icmltitlerunning{Latent Action Representation Alignment}

\begin{document}

\twocolumn[
  \icmltitle{LARA: Latent Action Representation Alignment for \\
  Vision-Language-Action Models
}



  \icmlsetsymbol{equal}{*}
  \icmlsetsymbol{lead}{$\dagger$}
  \begin{icmlauthorlist}
    \icmlauthor{Mengya Liu}{1,equal}
    \icmlauthor{Baoxiong Jia}{1,equal,lead}
    \icmlauthor{Jiangyong Huang}{1,2}
    \icmlauthor{Jingze Zhang}{1,2}
    \icmlauthor{Siyuan Huang}{1,3}
  \end{icmlauthorlist}
\centering{
\href{https://lmy1001.github.io/ICML26_LARA/}{https://lmy1001.github.io/ICML26\_LARA}}
  \icmlaffiliation{1}{State Key Laboratory of General Artificial Intelligence, BIGAI}
  \icmlaffiliation{2}{Peking University}
  \icmlaffiliation{3}{Delta Intelligence}

  \icmlcorrespondingauthor{Baoxiong Jia}  {baoxiongjia@g.ucla.edu}

  \icmlkeywords{Machine Learning, ICML}

  \begin{center}
    \parbox{\linewidth}{
      \centering
      \captionsetup{type=figure}
      \includegraphics[width=\linewidth]{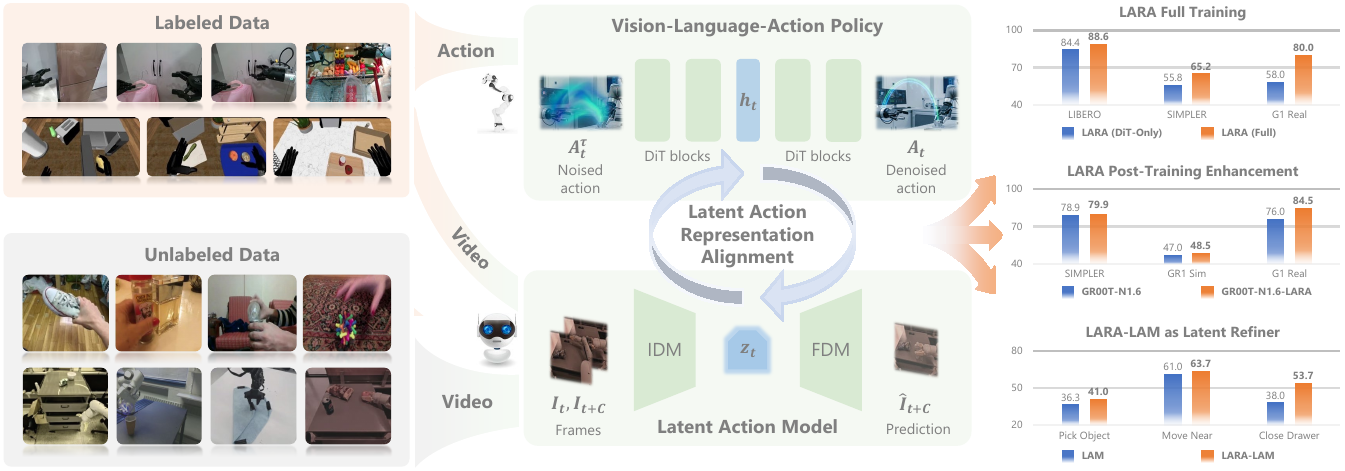}
      \captionof{figure}{We present \textbf{L}atent \textbf{A}ction \textbf{R}epresentation \textbf{A}lignment (LARA), a simple yet highly effective Vision-Language-Action (VLA) framework that bridges unlabeled video data and action-labeled robot datasets by jointly training a Latent Action Model (LAM) and a diffusion-based VLA model via latent action representation alignment. LARA supports versatile usage as a pre-training method, a post-training enhancement module for pre-trained VLA models, and a latent action refiner for LAM-based VLA models.}
      \label{fig:teaser}
    }
  \end{center}%
]


\printAffiliationsAndNotice{\icmlEqualContribution $^\dagger$Project lead}  

\begin{abstract}
Visual-language action (VLA) models enable robots to predict actions directly from observations and language instructions, but their performance depends on large-scale, high-quality data and is limited by the scarcity of real-world robot action datasets. To facilitate VLA model learning with abundant unlabeled human videos, Latent Action Models (LAM) learn latent action representations from visual dynamics to provide additional supervision for VLA learning. However, LAM and VLA are typically trained separately, leaving LAM ungrounded during VLA training and VLA models constrained by frozen LAM representations. To address these issues, we propose Latent Action Representation Alignment (LARA), a plug-and-play framework that jointly optimizes LAM and VLA via representation alignment. This enables reciprocal benefits where LAMs learn with action trajectories to avoid spurious visual changes, while VLAs are regularized by forward dynamics learned within LAMs to reduce hallucinations of functionally ineffective trajectories. We demonstrate LARA's versatility and effectiveness for pre-training, post-training enhancement of pre-trained VLA models, and LAM refinement, achieving an average of $\sim$10\%, $\sim$5\%, and $\sim$15\% improvement over 3 simulation and 1 meticulously designed real-world robotic manipulation benchmarks. The code is publicly available at \href{https://github.com/lmy1001/LARA}{https://github.com/lmy1001/LARA}.

\end{abstract}

\input{Sections/intro}
\input{Sections/related_work}
\input{Sections/preliminary}

\input{Sections/method}
\input{Sections/experiment}
\input{Sections/conclusion}



\section*{Impact Statement}
This paper presents work whose goal is to advance the field of Machine Learning. There are many potential societal consequences of our work, none of which we feel must be specifically highlighted here.

\bibliography{reference}
\bibliographystyle{icml2026}

\clearpage
\input{Sections/appendix}

\end{document}

%% file: configs.tex
\usepackage[dvipsnames]{xcolor}
\definecolor{Gray}{gray}{0.9}
\definecolor{mygreen}{rgb}{0.0, 0.5, 0.0}
\definecolor{myred}{rgb}{0.8, 0.25, 0.33}
\definecolor{myblue}{rgb}{0.19, 0.55, 0.91}
\definecolor{uclablue}{rgb}{0.15, 0.45, 0.68}
\definecolor{ucladblue}{rgb}{0.0, 0.33, 0.53}
\definecolor{ucladdblue}{rgb}{0.0, 0.23, 0.36}
\definecolor{uclagold}{rgb}{1.0, 0.82, 0.0}
\definecolor{ucladgold}{rgb}{1.0, 0.78, 0.17}
\definecolor{ucladdgold}{rgb}{1.0, 0.72, 0.11}
\definecolor{boxgreen}{rgb}{0.02, 0.66, 0.02}
\definecolor{boxred}{rgb}{0.66, 0.1, 0.1}
\definecolor{boxblue}{rgb}{0.01, 0.01, 0.73}

\usepackage{etoc}
\usepackage{lipsum}
\usepackage{wrapfig}
\usepackage{multicol}
\usepackage{mdframed}

\usepackage{mathtools}
\usepackage{amsthm}

\usepackage[utf8]{inputenc}   %
\usepackage[T1]{fontenc}      %
\usepackage{url}              %
\usepackage{amsfonts}         %
\usepackage{nicefrac}         %
\usepackage{microtype}
\usepackage{times}
\usepackage{graphicx}
\usepackage{amsmath,amssymb,mathbbol}
\usepackage{acronym}
\usepackage{enumitem}
\usepackage{balance}
\usepackage{xspace}
\usepackage{setspace}
\usepackage[skip=3pt,font=small]{caption}
\usepackage{booktabs,tabularx,colortbl,multirow,array,makecell}
\usepackage{pifont}
\usepackage{xr-hyper}
\usepackage{tcolorbox}
\tcbuselibrary{breakable}
\usepackage{newtxmath}
\usepackage{dsfont}

\usepackage[pagebackref,breaklinks,citecolor=ucladblue,colorlinks=true,linkcolor=ucladblue,bookmarks=false]{hyperref}
\usepackage[capitalize,noabbrev]{cleveref}

\usepackage{siunitx}   %
\usepackage{soul}   %
\usepackage{listings}
\definecolor{codegreen}{rgb}{0,0.6,0}
\definecolor{codegray}{rgb}{0.5,0.5,0.5}
\definecolor{codepurple}{rgb}{0.58,0,0.82}
\definecolor{backcolour}{rgb}{0.95,0.95,0.92}

\lstdefinestyle{mystyle}{
    backgroundcolor=\color{backcolour},   
    commentstyle=\color{codegreen},
    keywordstyle=\color{magenta},
    numberstyle=\tiny\color{codegray},
    stringstyle=\color{codepurple},
    basicstyle=\ttfamily\footnotesize,
    breakatwhitespace=false,         
    breaklines=true,                 
    captionpos=b,                    
    keepspaces=true,                 
    numbers=left,                    
    numbersep=5pt,                  
    showspaces=false,                
    showstringspaces=false,
    showtabs=false,                  
    tabsize=2
}

\makeatletter
\DeclareRobustCommand\onedot{\futurelet\@let@token\@onedot}
\def\@onedot{\ifx\@let@token.\else.\null\fi\xspace}
\def\eg{\emph{e.g}\onedot} 

\def\ie{\emph{i.e}\onedot}

\def\etc{\emph{etc}\onedot}

\makeatother

\makeatletter
\renewcommand{\paragraph}{%
  \@startsection{paragraph}{4}%
  {\z@}{0ex \@plus 0ex \@minus 0ex}{-1em}%
  {\hskip\parindent\normalfont\normalsize\bfseries}%
}
\makeatother

\crefname{algorithm}{Alg.}{Algs.}
\Crefname{algocf}{Algorithm}{Algorithms}
\crefname{section}{Sec.}{Secs.}
\Crefname{section}{Section}{Sections}
\crefname{table}{Tab.}{Tabs.}
\Crefname{table}{Table}{Tables}
\crefname{figure}{Fig.}{Fig.}
\Crefname{figure}{Figure}{Figure}
\crefname{equation}{Eq.}{Eq.}
\Crefname{equation}{Equation}{Equation}

\definecolor{gblue}{HTML}{4285F4}
\definecolor{gred}{HTML}{DB4437}
\definecolor{ggreen}{HTML}{0F9D58}

\definecolor{mygray}{gray}{.92}
\definecolor{up}{RGB}{34,139,34}   
\definecolor{down}{RGB}{178,34,34}        

\newcommand{\sota}{state-of-the-art\xspace}

\acrodef{lara}[LARA]{Latent Action Representation Alignment}
\acrodef{lam}[LAM]{Latent Action Model}
\acrodef{repa}[REPA]{Representation Alignment}
\acrodef{llm}[LLM]{Large Language Model}
\acrodef{vla}[VLA]{Vision-Language-Action}
\acrodef{vl}[VL]{Vision-Language}
\acrodef{vlm}[VLM]{Vision-Language Model}
\acrodef{rl}[RL]{Reinforcement Learning}
\acrodef{sft}[SFT]{Supervised Fine-Tuning}
\acrodef{nlp}[NLP]{Natural Language Processing}
\acrodef{sft}[SFT]{Supervised Fine-Tuning}
\acrodef{dit}[DiT]{Diffusion Transformer}
\acrodef{vanilla}[vanilla-VLA]{vanilla-VLA}
\acrodef{idm}[IDM]{Inverse Dynamic Model}
\acrodef{fdm}[FDM]{Forward Dynamic Model}
\acrodef{vae}[VAE]{Variational AutoEncoder}
\acrodef{ldm}[LDM]{Latent Diffusion Model}

\definecolor{line1}{rgb}{0.99,0.96,0.96}
\definecolor{line2}{rgb}{0.96,0.99,0.94}
\definecolor{line3}{rgb}{0.92,0.96,1.0}

%% file: math_commands.tex
\usepackage{amsmath,amsfonts,bm}

\def\eqref#1{equation~\ref{#1}}

\def\1{\bm{1}}

\def\rva{{\mathbf{a}}}

\def\rvc{{\mathbf{c}}}

\def\rvf{{\mathbf{f}}}

\def\rvh{{\mathbf{h}}}

\def\rvs{{\mathbf{s}}}

\def\rvv{{\mathbf{v}}}

\def\rvy{{\mathbf{y}}}
\def\rvz{{\mathbf{z}}}

\def\mA{{\bm{A}}}

\def\mI{{\bm{I}}}

\def\mL{{\bm{L}}}

\DeclareMathAlphabet{\mathsfit}{\encodingdefault}{\sfdefault}{m}{sl}
\SetMathAlphabet{\mathsfit}{bold}{\encodingdefault}{\sfdefault}{bx}{n}

\def\gL{{\mathcal{L}}}

\def\gN{{\mathcal{N}}}

\newcommand{\E}{\mathbb{E}}



%% file: Sections/intro.tex
\section{Introduction}
\label{sec:intro}
With the rise of Large \acp{vlm}, robotic manipulation is shifting from classical task planning and control to learning-based \ac{vla} models~\cite{kim2024openvla, bjorck2025gr00t, black2024pi_0, brohan2022rt} where actions are predicted directly given visual observation and language instruction. Like other \acp{vlm}, \ac{vla} performance critically depends on large-scale high-quality data. However, unlike visual datasets, robotic data is difficult and costly to collect, requiring real-world robot interactions, and is hard to generalize across different embodiments. Despite recent efforts to unify and scale robotic datasets~\cite{o2024open,fang2023rh20t,bu2025agibot}, robotic data remains scarce, causing even \sota \ac{vla} models to overfit and struggle to generalize to novel tasks and environments.

To overcome the robotic data bottleneck, human videos~\cite{goyal2017something,grauman2024ego} have been used as a rich data source due to their scale, accessibility, and diverse task coverage. However, the lack of robot action labels and the large human-robot embodiment gap prevent the direct use of human videos in robot learning. \acp{lam}~\cite{chen2022lapo,ye2024latent,chen2024moto} address this challenge by learning to predict future states and compressing visual dynamics into latent action representations as additional \ac{vla} data sources. This mechanism is either integrated into \ac{vla} models as a pre-training stage before action learning~\cite{bu2025univla} or learned separately to generate pseudo-labels for \ac{vla} learning~\cite{ye2024latent,chen2024moto}. In both cases, training involves \textit{complex, multi-stage training pipelines with model-specific designs}. More importantly, as shown in~\cref{fig:diff_lam_usage} (left), \ac{lam} learning is largely \textit{decoupled} from \ac{vla} learning, leaving \ac{lam} ungrounded on accurate action trajectories available during \ac{vla} learning and \ac{vla} models constrained by frozen \ac{lam} representations.

To this end, we propose \textbf{L}atent \textbf{A}ction \textbf{R}epresentation \textbf{A}lignment (LARA), a framework that bridges \ac{lam} and \ac{vla} learning via representation alignment (\cref{fig:diff_lam_usage} (right)) with the following key insights: 
\begin{itemize}[leftmargin=*,noitemsep,nolistsep,topsep=0pt,partopsep=0pt]
\item For \ac{lam}, joint learning with \ac{vla} models action trajectories grounds inverse visual dynamics learned to real actions, reducing the learning of spurious visual changes (\eg, background, lighting, \etc) from reconstruction.
\item For \ac{vla}, \ac{lam} regularizes learning by incorporating forward predictions of action effects, reducing hallucinations of kinematically plausible yet functionally incorrect or task-irrelevant action trajectories.
\end{itemize}
Drawing inspiration from recent work on representation alignment in diffusion models~\cite{yu2024representation,leng2025repa}, LARA coordinates \ac{lam} and \ac{vla} models through a lightweight mechanism compatible with most diffusion-based \ac{vla} architectures. Through extensive experiments on simulation and real-world robotic benchmarks, we demonstrate LARA as: (1) \textbf{a strong \ac{vla} training pipeline}, improving base \ac{vla} models by $\sim$10\%, (2) \textbf{a powerful post-training enhancement module} for pre-trained \acp{vla}, yielding $\sim$5\% improvement on average, and (3) \textbf{an effective method for refining latent action representations} in \ac{lam}, boosting based model performance by $\sim$15\% when used as pseudo-labels for downstream \ac{vla} learning. Our contributions are as follows:

\begin{figure}[t!]
    \centering
    \includegraphics[width=1.0\linewidth]{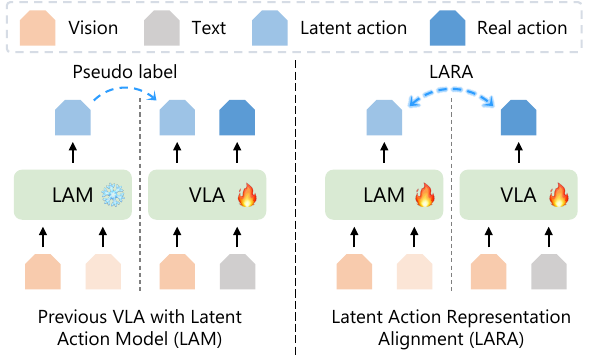}
    \vspace{-10pt}
    \caption{\textbf{Comparison of \ac{lam}-based \ac{vla} models.} \acp{lam} are commonly used as pseudo labels for \ac{vla} learning (left), where as LARA jointly optimizes \ac{lam} and \ac{vla} model by explicitly aligning their latent representations (right).}
    \label{fig:diff_lam_usage}
    \vspace{-17pt}
\end{figure}

\begin{itemize}[leftmargin=*,noitemsep,nolistsep,topsep=0pt,partopsep=0pt]
    \item We propose, LARA, a novel and effective framework for jointly improving \ac{lam} and \ac{vla} model learning via latent action representation alignment.
    \item We show LARA's versatility as a pre-training method, a plug-and-play post-training enhancement module, and a latent action refiner for \acp{lam}.
    \item We validate LARA on 3 challenging simulation and 1 real-world robot benchmarks, achieving $\sim$10\%, $\sim$5\%, and $\sim$15\% improvements for full training, post-training enhancement, and \ac{lam} refinement, respectively.
\end{itemize}

%% file: Sections/related_work.tex
\section{Related Works}
\label{sec:related_work}
\paragraph{Vision-Language-Action (VLA) Models.} 
VLA models leverage the reasoning capabilities of \acp{vlm}~\cite{karamcheti2024prismatic, li2025eagle, huang2024embodied, gong2023arnold} to integrate natural language instructions, visual observations, and robot proprioception into a unified control policy. Pioneering works such as RT-1~\cite{brohan2022rt} and Octo~\cite{team2024octo} employ a transformer-based
policy that integrates diverse data, including robot trajectories across various tasks. RT-2~\cite{brohan2024rt}, OpenVLA~\cite{kim2024openvla},  $\pi_0$~\cite{black2024pi_0}, and GR00T-N1~\cite{bjorck2025gr00t} adopt a paradigm of large-scale cross-embodiment pre-training followed by task-specific fine-tuning, achieving strong inference performance. Subsequent studies further enhance these models along multiple dimensions, including spatial reasoning~\cite{qu2025spatialvla, zhang2025dig, zhen20243d, li2026pointvla}, long-horizon planning via Chain-of-Thought~\cite{lin2025onetwovla,zhao2025cot}, and hierarchical policy systems~\cite{luo2025being,shi2025hi, huang2025thinkact, lee2025molmoact, li2025cogvla}. Despite these advances, these models are still limited by the scarcity and high cost of labeled robot data~\cite{o2024open} compared to the vast availability of unlabeled video data, revealing a largely untapped opportunity to leverage motion priors inherently embedded in general video corpora.

\paragraph{Latent Action Models (LAM) for VLA Pretraining.} Latent action learning originated in general video domains with approaches such as LAPO~\cite{chen2022lapo} and Genie~\cite{bruce2024genie}, inferred latent control signals to model video dynamics. In robotics, \ac{lam} is commonly used as an intermediate representation to supervise \ac{vla} learning~\cite{ye2024latent,chen2024moto, bjorck2025gr00t}. The subsequent study on \ac{lam} improves this paradigm in terms of latent action quality~\cite{nikulin2025latent, liang2025clam}, data scaling~\cite{chen2025villa}, and better integration within \ac{vla} models~\cite{bu2025univla}. However, these methods generally treat the \ac{lam} as a static provider of pseudo-labels or pre-trained weights. This decoupling prevents latent representations from adapting to real robot actions, leaving the gap between visual dynamics and robot motor execution unresolved.

\paragraph{Representation Alignment.} The idea of leveraging good representations for regularizing modeling has proven to be effective for a diverse range of tasks, \ac{vlm} learning~\cite{jain2025elevating}, 3D understanding~\cite{huang20253drs}, as well as image generation~\cite{ye2024latent,leng2025repa, ma2025unitok, yao2025denoising}. Despite similar ideas having been explored on \ac{vla} learning~\cite{zheng2025flare,kachaev2025don} by aligning the \ac{vla} model with diverse frozen visual-language features, we argue that the alignment target should essentially be an updatable action representation to allow the latent action space to co-evolve with \ac{vla} learning. In fact, we show that, similar to REPA~\cite{ye2024latent} in image generation, this can be easily achieved by a bidirectional representation alignment loss between intermediate features of the diffusion \ac{vla} and \ac{lam} latent actions.

%% file: Sections/preliminary.tex
\section{Background}
\label{sec:prelim}
\paragraph {Diffusion-based \ac{vla} Models}
\label{sec:prelim:vla_dit}
Flow-based \ac{vla} models map visual observations and natural language instructions to robot action trajectories using \acp{vlm} and flow-based generative models. At timestep $t$, the \ac{vlm} model $\texttt{VLM}_{\theta}$ extracts task-related vision-language tokens $\rvf_{t}^{\text{vl}} = \texttt{VLM}_{\theta}(\mI_t, \mL)$ from the instruction $\mL$ and observation $\mI_t$. Combined with robot proprioceptive state $\rvs_t$, these form the conditioning input $\rvc_t = \{\rvs_t, \rvf_{t}^{\text{vl}}\}$ for action generation. The diffusion-based action generation model then generates an action chunk with $C$ steps $\mA_{t} = \rva_{t:t+C}$ via flow matching. Specifically, given the flow timestep $\tau\in[0,1]$ and the sampling noise $\bm{\epsilon}\sim\gN(\bm{0},\mI)$, the model optimizes:
\begin{equation}
\label{eq:action_loss}
    \begin{aligned}
    \mathcal{L}_{\text{ACT}}(\theta) = \mathbb{E}_{\tau, \bm{\epsilon}} \left[ \| v_\theta(\mA_t^{\tau} , \rvc_t) - (\mA_t - \bm{\epsilon}))\|^2 \right],
    \end{aligned}
\end{equation}
where $\mA_t^{\tau} = \tau\mA_t + (1 - \tau)\bm{\epsilon}$ is the noised action. This objective trains the velocity field network $v_\theta$ to predict denoising directions at each flow timestep~\cite{lipman2022flow}. With this velocity field, we can generate actions from random noise $\mA_t^0\sim\gN(\bm{0},\mI)$ by integrating $v_\theta$ from $\tau=0$ to $\tau=1$ via the forward Euler rule:
\begin{equation}
   \mA_t^{\tau+\frac{1}{K}}= \mA_t^{\tau} + \frac{1}{K} v_\theta(\mA^{\tau}, \rvc_t),
\end{equation}
where $K$ is the number of integration steps controlling the approximation accuracy. 

\paragraph{Latent Action Model (LAM)} Given the scarcity of action-labeled robotic data, prior works have explored leveraging unlabeled human and robot interaction videos for robotic action learning~\cite{chen2022lapo,ye2024latent,chen2024moto}. These methods employ a \ac{lam} that encodes transitions between current visual observations $\mI_t$ and future observation $\mI_{t+C}$ into discrete latent actions. Specifically, the \ac{lam} consists of three components: 
\noindent\begin{enumerate}[leftmargin=*,noitemsep,nolistsep]
\item[(1)] \textit{\ac{idm}}  $\rvz_t=\texttt{IDM}_\varphi(\mI_{t}, \mI_{t+C})$ predicts continuous latent action $\rvz_t$ capturing implicit dynamics between current and future observations.
\item[(2)] \textit{Vector Quantizer} $\rvz_{t}^q=\texttt{Quant}_\varphi(\rvz_t)$, that discretizes the latent into a codebook token $\rvz_t^q\in\{\rvz_\varphi^1, \cdots, \rvz_\varphi^K\}_{\text{codebook}}$ following VQ-VAE~\cite{van2017neural}.
\item[(3)] \textit{\ac{fdm}} $\hat{\mI}_{t+C} = \texttt{FDM}_\varphi(\mI_t,\rvz_t^q)$, that reconstructs the future observation conditioned on the current observation and the quantized latent action.
\end{enumerate}
The full pipeline is trained end-to-end with the VQ-VAE~\cite{van2017neural} objective:
\begin{equation}
    \label{eq:lam_loss}
    \small
    \gL_{\text{LAM}}(\varphi) = \|\mI_{t+C} - \hat{\mI}_{t+C}\|_2^2 + \|\text{sg}[\rvz_t^q] - \rvz_t\|_2^2 + \beta\|\rvz_t^q - \text{sg}[\rvz_t]\|_2^2,
\end{equation}
where $\text{sg}[\cdot]$ denotes the stop-gradient operation and $\beta$ balances the commitment loss~\cite{van2017neural}. Most \ac{lam}-based \ac{vla} models~\cite{ye2024latent,chen2024moto,chen2025villa} follow a two-stage protocol: first pre-training the \ac{lam} on unlabeled data, then leveraging it to guide \ac{vla} training with labeled data. As illustrated in~\cref{fig:diff_lam_usage} (left), the standard approach treats latent action tokens $\rvz^q_t$ as additional supervision, where the \ac{vla} model is trained to predict both the latent action $\rvz_q^t$ and the actual low-level actions $\mA_t$. While this design facilitates model training, it potentially risks constraining the quality of the learned action representations to the fidelity of the pseudo labels produced by the \ac{lam} given similar limitation observed in the image generation domain~\cite{leng2025repa}.

%% file: Sections/method.tex
\section{Method}
\label{sec:method}

\begin{figure*}[t!]
    \centering
    \includegraphics[width=\linewidth]{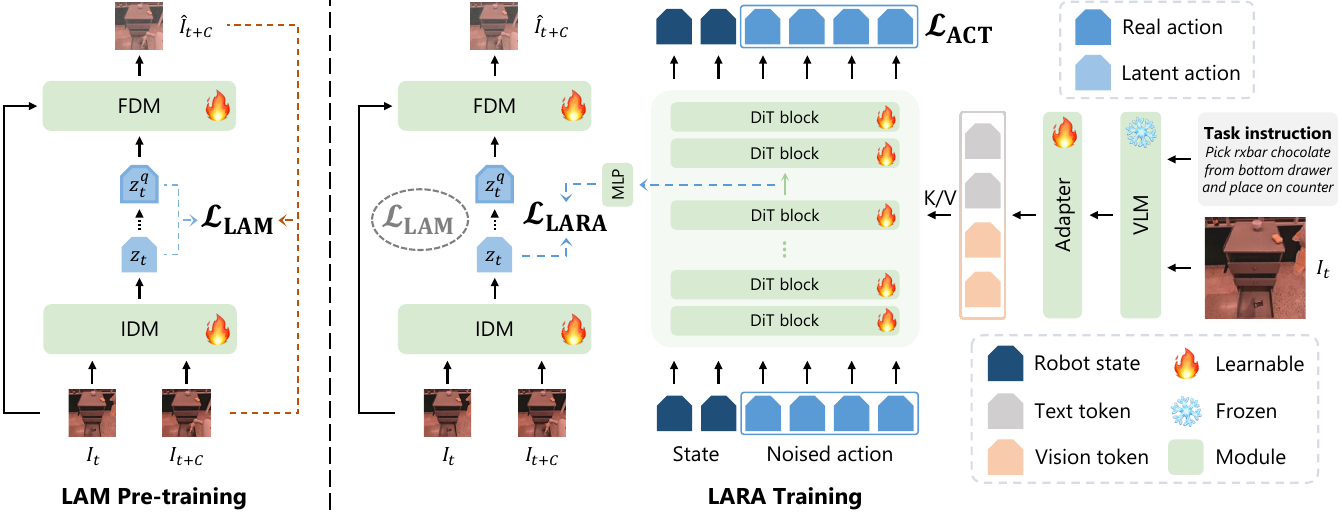}
    \caption{\textbf{Method overview.} We begin with \ac{lam} (left), where an \acf{idm} learns a latent action $
    \rvz_t$ from consecutive image frames, and a \acf{fdm} learns to reconstruct the subsequent frame conditioned on the preceding frame and the quantized latent action $\rvz_t^q$. We then conduct \ac{lara} training on a diffusion-based \ac{vla} model, where \ac{lara} explicitly aligns the latent action $\rvz_t$ with intermediate features of the \ac{dit}, thereby jointly optimizing the \ac{lam} and \ac{vla} model in an end-to-end manner.}
    \label{fig:method_overview}
    \vspace{-15pt}
\end{figure*}

As discussed in~\cref{sec:intro} and~\cref{sec:prelim}, \ac{lam} and flow-based action models represent complementary aspects of robot control but operate in isolation without leveraging each other's modeling on state transition (effect) and action commands (cause). To this end, we propose LARA (\textbf{L}atent \textbf{A}ction \textbf{R}epresentation 
\textbf{A}lignment) to enable joint optimization of both models via a simple and effective latent action representation alignment mechanism. We provide an overview of LARA in~\cref{fig:method_overview}.

\subsection{Latent Action Representation Alignment (LARA)}
\label{sec:method:lara}

\paragraph{Latent Action Representation Alignment} Recent work in image generation~\cite{yu2024representation,leng2025repa}, has demonstrated that aligning diffusion model intermediate features with pre-trained representations like DINOv2~\cite{oquab2023dinov2} improves generation quality. We adopt this principle for action generation by treating the flow-matching model $v_\theta(\mA_t^\tau, \rvc_t)$ from~\cref{eq:action_loss} as an encoder-decoder structure $E_\theta\circ D_\theta$ within its architecture:
\begin{equation}
    \rvh_t^\theta = E_\theta(\mA_{t}^\tau, \rvc_t),\quad \hat{\rvv}_t = D_\theta(\rvh_t^\theta, \rvc_t),
\end{equation}
where the encoder $E_\theta$ extracts an intermediate latent representation $\rvh_t^\theta$, which the decoder $D_\theta$ uses to predict the target velocity $\rvv_t = \mA_t - \bm{\epsilon}$. In practice, when implemented as a \ac{dit}~\cite{peebles2023scalable}, the representation $\rvh_t^\theta$ corresponds to the latent features between \ac{dit} layers. Drawing inspiration from image diffusion methods like REPA~\cite{yu2024representation}, we introduce a representation alignment objective for action learning. Given a frozen pre-trained action representation embedding $\rvy_{t}^{\text{pretrain}}$, we optimize:
\begin{equation}
    \label{eq:repa}
    \small
    \gL_{\text{RA}}(\theta, \psi) = -\E_{\mA_t, \bm{\epsilon}, \tau}\left[\texttt{CosSim}\left(\rvy_{t}^{\text{pretrain}}, f_\psi(\rvh_{t}^\theta)\right)\right],
\end{equation}
where $f_\psi(\cdot)$ is a learnable projection head that adapts between the pre-trained action representation space and the diffusion feature space. Different from prior works~\cite{zheng2025flare} that leverage frozen action embeddings, we leverage the online \ac{lam} latent actions $\rvz_t^{\text{frozen}}$ from~\cref{eq:lam_loss} as $\rvy_t^{\text{pretrain}}$ and propose the joint training of the \ac{lam} and action diffusion models. Specifically, we replace the frozen representation $\rvy_t^{\text{pretrain}}$ with the online \ac{lam} latent action:
\begin{equation}
\label{eq:lara}
    \small
    \gL_{\text{LARA}}(\theta, \varphi, \psi) = -\E_{\mA_t, \bm{\epsilon},\tau}\left[\texttt{CosSim}\left(\rvz_t^\varphi, f_\psi(h_t^\theta)\right)\right],
\end{equation}
where $\rvz_t^\varphi = \rvz_t$ is the online continuous latent action before quantization in the \ac{lam}. This alignment loss is combined with both the flow-matching objective and the \ac{lam} reconstruction objective to form the full LARA objective:
\begin{equation}
\label{eq:full_objective}
\small
\mathcal{L}(\theta,\varphi,\psi) = \mathcal{L}_{\text{ACT}}(\theta) + w_1\mathcal{L}_{\text{LARA}}(\theta,\varphi,\psi) + w_2\mathcal{L}_{\text{LAM}}(\varphi),
\end{equation}
where $w_1$ and $w_2$ are loss balancing hyperparameters.

\paragraph{Bi-directional Regularization Effect} Notably, LARA induces complementary regularization effects on both the \ac{lam} and the action diffusion model:
\begin{enumerate}[leftmargin=*,nolistsep,noitemsep]
\item[(1)] \textit{Inverse Dynamics Regularization for \ac{lam}}: By aligning \ac{lam} latent actions with action policy representations, we constrain the action latent space to emphasize control-relevant visual changes rather than nuisance variations (\eg, lighting, shadows) which are irrelevant for action execution. This alignment suppresses these spurious factors arising from purely visual dynamics learning and encourages $\rvz_t$ to encode only causal features necessary for predicting actions, resulting in a more action-centric \ac{lam} latent space.
\item[(2)] \textit{Forward Dynamics Grounding for Action Diffusion}: Standard behavior cloning-based action generation largely reduces to pattern matching from observations to actions, without explicitly modeling the physical consequences of actions. By anchoring intermediate action-\ac{dit} representations to the forward-predictive latent actions learned by the \ac{lam}, we inject an explicit notion of future state evolution into the action diffusion policy. This grounding biases the action model toward representations consistent with plausible future world states, mitigating the prediction of kinematically hallucinated trajectories (\ie, physically plausible but non-effect actions), and ensuring that generated actions respect environment dynamics.
\end{enumerate}

\subsection{Training and Application of LARA}
\label{sec:method:application}
\paragraph{Model Design} For the \ac{lam} model, we adopt the latent action model design from Moto-GPT~\cite{chen2024moto}, modeling the \texttt{IDM} and \texttt{FDM} using ViT-based encoder-decoder architectures and a latent codebook size of 128. For the diffusion \ac{vla} model, we use a standard cross-attention \ac{dit} backbone following prior works~\cite{liu2024rdt,bjorck2025gr00t}. Vision-language features are extracted using a frozen Eagle-2~\cite{li2025eagle} \ac{vlm} with learnable adapters. To accommodate for diverse robot embodiments, we follow GR00T-N1~\cite{bjorck2025gr00t} and employ embodiment-specific MLP encoders that map heterogeneous proprioceptive state sapces into a shared embedding space before feeding them into the \ac{dit}. LARA at the second-to-last $L-2$ layer of the \ac{dit}. We provide additional implementation and training details in~\cref{app:details}.
 

\paragraph{LARA Training Pipeline} We train both the \ac{lam} and the action diffusion models with the following three stages:

\begin{enumerate}[leftmargin=*,nolistsep,noitemsep]
\item[(1)] \textit{\ac{lam} Pre-training}: We train the \ac{lam} on large-scale unlabeled video data, including both robot data and internet videos using the \ac{lam} objective as in~\cref{eq:lam_loss}. This establishes a general purpose latent action space capturing visual dynamics across diverse scenarios.
\item[(2)] \textit{LARA Joint Pre-Training}: Taking the reconstruction pre-trained \ac{lam} from stage 1, we train the action diffusion model on robot demonstration data with action labels, applying the full LARA objective in~\cref{eq:full_objective} and updating the action \ac{dit} and \ac{lam} jointly. This stage incorporates diverse robot embodiments to learn better embodiment-agnostic action representations.
\item[(3)] \textit{LARA Joint Post-Training}: We fine-tune all models from stage 2 on target-task demonstrations for the deployment embodiment with task-specific data. This follows the standard \ac{vla} pre-training and post-training paradigm.
\end{enumerate}

\paragraph{Applications of LARA} As the representation-level alignment in LARA enables flexible integration with diverse pre-trained \acp{lam} and action diffusion models without architectural modifications, we demonstrate two representative applications of LARA on pre-trained models:
\begin{enumerate}[leftmargin=*,noitemsep,nolistsep]
\item[(1)] \textit{LARA Post-training Enhancement}, where LARA is applied as a modular post-training procedure to an existing pre-trained diffusion-based \ac{vla} model using a pretrained \ac{lam} for representation alignment.

\item[(2)] \textit{LARA for Latent Action Refinement}, where the LARA-pretrained \ac{lam} provides improved structured latent action tokens that can be directly used as pseudo-labels in latent action-based frameworks such as LAPA~\cite{ye2024latent} and Moto-GPT~\cite{chen2024moto}.
\end{enumerate}

Both usage modes rely on post-training-scale data and are empirically evaluated in~\cref{sec:exp} to show their effectiveness.

%% file: Sections/experiment.tex
\section{Experiments}
\label{sec:exp}
In this section, we validate the efficacy of \ac{lara} by addressing the research questions below in the following sections:
\begin{enumerate}[leftmargin=*,noitemsep,nolistsep]
    \item[(1)] How does LARA compare to existing \ac{vla} models across diverse robotic menchmarks?
    \item[(2)] To what extent does LARA improve existing models as a post-training refinement module for \acp{vla} and as a latent action refiner for \acp{lam}?
    \item[(3)] How well does LARA generalize to novel tasks and robot embodiments, and which factors are critical for effective LARA training?

\end{enumerate}

\begin{table*}[t!]
    \caption{\textbf{Benchmark Evaluations}. We show the performance of LARA variants against existing models under the \textit{OXE-Constrained} and \textit{Unconstrained} settings. For GR00T-N1.6-LARA, we post-train GR00T-N1.6 with an OXE-pretrained \ac{lam} using LARA.}
    \centering
    \resizebox{0.97\linewidth}{!}{%
    \begin{tabular}{l c c c c  c c c c c c c}
    \toprule
    \multirow{2}[2]{*}{Methods} & \multicolumn{5}{c}{\textit{LIBERO}} & \multicolumn{4}{c}{\textit{SIMPLER-ENV}} \\
    \cmidrule(lr){2-6}\cmidrule(lr){7-10}
    & Spatial & Object & Goal & Long & Average & Pick & Move & Drawer & Average \\
       \midrule
       \rowcolor[HTML]{F0F7FF}\multicolumn{10}{l}{\footnotesize{\textit{OXE-Constrained Comparison}}}\\
        OpenVLA~\cite{kim2024openvla} & 84.7 &	88.4 &	79.2 &	53.7 &	76.5 & 16.3 & 46.2 & 35.6 &32.7\\
        Octo~\cite{team2024octo}  & 78.9 &	85.7 	& 84.6 & 	51.1 &	75.1 &   17.0 & 4.2 & 22.7 &  14.6 \\
       Moto-GPT~\cite{chen2024moto} & - & -& - & - & - & 74.0 & 60.4 & \textbf{43.1} & 61.4  \\
       LAPA~\cite{ye2024latent} & 73.8 &74.6  &58.8 & 55.4 & 65.7 & - & - & - & - \\
       \midrule
       LARA (\textit{DiT-only})  & 84.5 & 90.0 & 86.5 & 76.5 & 84.4 &  62.3 & \textbf{84.0} & 21.0 & 55.8 \\
       LARA (\textit{full}) & \textbf{88.0} & \textbf{92.0} & \textbf{88.5} & \textbf{86.0} & \textbf{88.6} & \textbf{82.3} & 83.7 & 29.5 & \textbf{65.2} \\
       \textit{LARA Improvement} & \textcolor{up}{+4.1\%} & \textcolor{up}{+2.2\%} & \textcolor{up}{+2.3\%} & \textcolor{up}{+12.4\%} & \textcolor{up}{+5.0\%}& \textcolor{up}{+32.1\%} & \textcolor{down}{-0.4\%} & \textcolor{up}{+40.5\%} & \textcolor{up}{+16.8\%} \\
      \hline
      \hline
      \rowcolor[HTML]{F0F7FF}\multicolumn{10}{l}{\footnotesize{\textit{Unconstrained Comparison}}}\\
       SpatialVLA~\cite{qu2025spatialvla} &	88.2 & 	89.9 &	78.6 & 	55.5 &	78.1 & 88.0 & 72.7 & 41.8 & 70.7\\
       CoT-VLA~\cite{zhao2025cot} & 87.5 & 91.6 & 87.6 & 69.0 & 81.1 & - & - & - & - \\
        $\pi$0-FAST~\cite{pertsch2025fast}& 96.4 & 96.8 & 88.6 & 60.2 & 85.5 & 75.3 & 67.5 &  42.9 & 61.9 \\
        UniVLA~\cite{bu2025univla}  & 96.5 & 96.8 & 95.6 & 92.0 & 95.2 & - & - & - & - \\
       TraceVLA~\cite{zheng2024tracevla} &- &- &- &- & -&  45.0 &63.8 & 63.1 & 57.3 \\ 
       Magma~\cite{yang2025magma} & -&- &- &- &- & 75.0 & 53.0 & 58.9& 62.3\\
       villa-X~\cite{chen2025villa} & \textbf{97.5} & 97.0 & 91.5 & 74.5 &  90.1 &  \textbf{98.7} & 75.0 & \textbf{59.3} & 77.7\\
       DreamVLA~\cite{zhang2025dreamvla}  & \textbf{97.5} & 94.0 & 89.5 & 89.5 & 92.6 & - & - & - & - \\
       \midrule

        GR00T-N1.6~\cite{bjorck2025gr00t} &  \textbf{97.5} &96.0 & 95.5 & 91.0 & 95.0 & 97.3 & 87.0 & 52.3 &  78.9\\
       GR00T-N1.6-LARA &  96.5 &  \textbf{97.5} & \textbf{96.0}  & \textbf{92.5} &   \textbf{95.6} & 98.0 & \textbf{89.0} & 52.8 & \textbf{79.9} \\
       \textit{LARA Post-train Improvement} & \textcolor{down}{-1.0\%} & \textcolor{up}{+1.6\%} & \textcolor{up}{+0.5\%} & \textcolor{up}{+1.6\%} & \textcolor{up}{+0.6\%} & \textcolor{up}{+0.7\%} & \textcolor{up}{+2.3\%} & \textcolor{up}{+1.0\%} & \textcolor{up}{+1.3\%} \\ 
       \bottomrule
    \end{tabular}
}
    \vspace{-10pt}
    \label{tab:benchmark_test}
\end{table*}

\subsection{Experimental Settings}
\label{sec:exp:settings}
\paragraph{General Experimental Settings} To ensure fair comparison with existing \ac{vla} models that vary widely in pre-training data scale and sources, we evaluate \textsc{Lara} under two distinct 
experimental settings.:
\begin{itemize}[leftmargin=*,noitemsep,nolistsep]
\item \textit{OXE-Constrained Comparison}: All models are pre-trained exclusively on datasets within the scope of Open-X-Embodiment (OXE)~\cite{o2024open} (see in~\cref{app:data_details}) and optionally post-trained on the target evaluation datasets. This setting offers a clean experimental setting to reveal the effect of specific model design.
\item \textit{Unconstrained Comparison}: We put no constraints on model design and pre-training datasets. This setting aims to reveal the true performance limit of designed models compared with \sota models.
\end{itemize}

\paragraph{Evaluation Setup} For evaluation, we assess model performance on three existing simulation benchmarks:
\begin{itemize}[leftmargin=*,noitemsep,nolistsep]
\item \textit{LIBERO}~\cite{liu2023libero}: We follow common post-training and evaluation protocols~\cite{bu2025univla} and report model success rates on \textit{LIBERO-Spatial}, \textit{LIBERO-Object}, \textit{LIBERO-Goal}, and \textit{LIBERO-Long}.
\item \textit{SIMPLER-ENV}~\cite{li2024evaluating}: We follow established post-training and evaluation protocols from villa-X~\cite{chen2025villa} and report model success rates on three task categories, including \textit{Pick Coke Can}, \textit{Object Movement}, and \textit{Open \& Close Drawer}.
\item \textit{GR1-Sim-24(30)}~\cite{bjorck2025gr00t}: We follow GR00T-N1~\cite{bjorck2025gr00t} and select the post-training setting with 30 demos for model training and report model success rates on the 24 tasks available.
\end{itemize}

\begin{figure*}[t!]
    \centering
    \begin{minipage}[c]{1.0\linewidth}
        \centering
    \captionof{table}{\textbf{Quantitative results on GR1 Simulation and G1 Real-World Evaluation.} We report success rates for the \textit{GR1-Sim-24(30)} benchmark (24 bimanual simulation tasks, fine-tuned on 30 demonstrations per task following~\citet{bjorck2025gr00t}) and the \textit{G1-Real(50)} suite. The real-world evaluation is conducted on the Unitree G1 humanoid across two multi-stage tasks (\textit{Pick-n-Place}, \textit{Grasp-n-Pour}), averaging performance over 50 trials per task.}
    \centering
    \resizebox{\linewidth}{!}{
    \begin{tabular}{l  c c c c c c c c c}
    \toprule
    \multirow{2}[2]{*}{Methods} & \multirow{2}[2]{*}{\makecell{\textit{GR1-Sim-24}\\Avg.}} & \multicolumn{3}{c}{\textit{G1-Real Pick-n-Place}} & \multicolumn{4}{c}{\textit{G1-Real Grasp-n-Pour}} & \multirow{2}[2]{*}{\makecell{\textit{G1-Real}\\Avg.}}\\ 
    \cmidrule(lr){3-5}\cmidrule(lr){6-9}
  & & Pick &  Place & Full &  Grasp-Left & Grasp-Right & Pour & Full & \\
        \midrule
        \rowcolor[HTML]{F0F7FF}\multicolumn{10}{l}{\footnotesize{\textit{OXE-Constrained Comparison}}}\\
         LARA (\textit{DiT-only}) &   6.4 & 74.0 & 78.4 &58.0 & 58.0 & 78.0 & 93.1 & 54.0 & 56.0\\
        \ac{lara} (\textit{full}) & 11.4  &  90.0 & 88.9 & 80.0 & 80.0 &   \textbf{84.0} &  \textbf{100.0} & \textbf{68.0} &  74.0\\
         \textit{LARA Improvement} & \textcolor{up}{+78.1\%} & \textcolor{up}{+21.6\%} & \textcolor{up}{+13.4\%} & \textcolor{up}{+37.9\%} & \textcolor{up}{+37.9\%} &  \textcolor{up}{+7.7\%} & \textcolor{up}{+7.4\%} & \textcolor{up}{+25.9\%} & \textcolor{up}{+32.1\%} \\
        \hline\hline
        \rowcolor[HTML]{F0F7FF}\multicolumn{10}{l}{\footnotesize{\textit{Unconstrained Comparison}}}\\
         GR00T-N1.6~\cite{bjorck2025gr00t} &  47.0 & 90.0 & 84.4 & 76.0 & 78.0 & 80.0 & 87.2 &   \textbf{68.0} & 72.0\\
        GR00T-N1.6-LARA & \textbf{48.5} & \textbf{92.0} & \textbf{91.3} & \textbf{84.0} &  \textbf{86.0} & 76.0 & 94.4 &  \textbf{68.0} &  \textbf{76.0} \\
        \textit{LARA Post-train Improvement} & \textcolor{up}{+3.2\%} & \textcolor{up}{+2.2\%} & \textcolor{up}{+8.2\%} & \textcolor{up}{+10.5\%} & \textcolor{up}{+10.3\%} & \textcolor{down}{-4.0\%} & \textcolor{up}{+7.2\%} & \textcolor{up}{+0.0\%} & \textcolor{up}{+5.56\%}\\
        \bottomrule
    \end{tabular}
    }
    \label{tab:gr1_g1_evaluation}
    \end{minipage}
    \begin{minipage}{\linewidth}
        \centering
        \vspace{0.3cm}
        \includegraphics[width=\linewidth]{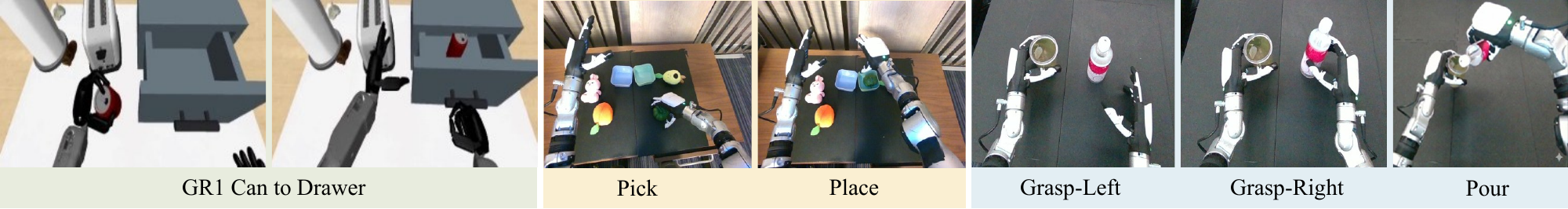}
        \caption{\textbf{Task Visualization of \textit{GR1-Sim-24(30)} and \textit{G1-Real(50)}.}We illustrate a representative bimanual task from the \textit{GR1-Sim-24(30)} simulation suite (left) alongside the two real-world tasks evaluated on the G1 humanoid: \textit{Pick-n-Place} and \textit{Grasp-an-Pour} (right). For a detailed frame-by-frame breakdown of the \textit{G1-Real(50)} execution, please refer to~\cref{fig:g1_data}.}
        \label{fig:task_visualization}
    \end{minipage}
    \vspace{-10pt}
\end{figure*}

Additionally, we meticulously design a real-world robot manipulation benchmark for model performance testing:
\begin{itemize}[leftmargin=*,noitemsep,nolistsep]
\item \textit{G1-Real(50)}: In \textit{G1-Real(50)}, we deploy models on a real-world Unitree G1 humanoid robot and test task performance on two composite tasks: (1) \textit{``Pick Green Tomate and Place in Green Basket''}, and (2) \textit{``Grasp Bottle and Pour to Cup''}. During post-training, we provide 50 real-robot manipulation demonstrations for each task. During evaluation, we model success rates on both sub-task (\eg, grasp first then pour) and full-task execution over 50 trials. We provide a visualization of this task in~\cref{fig:task_visualization} and additional details for real-robot setup in~\cref{app:g1_exp}.
\end{itemize}

\subsection{LARA for Full \ac{vla} Training}
\label{sec:exp:lara_vla}

\paragraph{Experimental Setup} We evaluate the efficacy of LARA as a full \ac{vla} framework, encompassing both pre-training and post-training stages under the \textit{OXE-constrained Comparison} setting. Specifically, we provide two LARA variants under this setting with training details in~\cref{app:details:training}:
\begin{itemize}[leftmargin=*,nolistsep,noitemsep]
\item LARA (\textit{DiT-only}): We pre-train a vanilla \ac{dit} model described in~\cref{sec:method:application} directly on OXE data without representation alignment and then post-train on target datasets.
\item LARA (\textit{full}): We train the full LARA model by first pre-training the \ac{lam} model on OXE data with only reconstruction loss. This LAM is used for LARA joint pre-training on OXE-data and post-training on target datasets with an \ac{dit} initialized from scratch.
\end{itemize}

\paragraph{Results \& Analyses} As detailed in the top section of~\cref{tab:benchmark_test}, LARA consistently outperforms existing \ac{vla} frameworks when pre-trained on OXE data, achieving a $12\%$ and $4\%$ overall improvements over the best baselines on \textit{LIBERO} and \textit{SIMPLER-ENV}, respectively. This includes \sota \ac{lam}-based \ac{vla} models like Moto-GPT~\cite{chen2024moto} and LAPA~\cite{ye2024latent}.  Additionally, compared to the vanilla \ac{dit} baseline, LARA (\text{\ac{dit}-only}), the full LARA pipeline reaches around $5\%$ and $15\%$ on LIBERO and SIMPLER, respectively, validating the effectiveness of LARA as a general and superior \ac{vla} model learning paradigm. Remarkably, despite using only OXE-data for pre-training, LARA outperforms several large-scale pre-trained models in the \textit{Unconstrained Comparison} setting, including $\pi_0$-FAST~\cite{pertsch2025fast}, Magma~\cite{yang2025magma}, and SpatialVLA~\cite{qu2025spatialvla}, demonstrating its potential as a data-efficient \ac{vla} framework.

\subsection{LARA for Post-training Enhancement}
\label{sec:exp:lara_post}
\paragraph{Experimental Setup} We evaluate LARA as a plug-and-play post-training enhancement module for existing diffusion-based \ac{vla} models under the \textit{Unconstrained Comparison} setting. Due to computational constraints, we select GR00T-N1.6~\cite{bjorck2025gr00t} as our baselines given their strong performance on a wide range of tasks. Specifically, we create the LARA enhanced model, GR00T-N1.6-LARA, by applying the full LARA objective to jointly train the pre-trained GR00T-N1.6 model with an \ac{lam} (pre-trained on OXE with reconstruction loss only) on the post-training data respectively. We provide  training details in~\cref{app:details:training}. In addition to the experiments conducted on GR00T-N1.6, we further apply LARA to the $\pi_{0.5}$~\cite{black2025pi_} model during post-training to validate its effectiveness across different backbone architectures. Additional experimental setups and results are provided in~\cref{app:pi0.5_exp}.

\paragraph{Results \& Analyses} As shown in the bottom section of~\cref{tab:benchmark_test} and bottom section of~\cref{tab:gr1_g1_evaluation}, our LARA enhanced model consistently outperforms existing models on all benchmarks. Notably, even when applied only in the post-training stage, adding LARA substantially improves performance over the vanilla GR00T-N1.6 model and the  $\pi_{0.5}$ model (see in ~\cref{tab:pi0.5_result}), achieving \sota performance. Compared to implicit world modeling models like DreamVLA~\cite{zhang2025dreamvla} and UniVLA~\cite{bu2025univla} that require full model re-training, LARA post-training enhancement is significantly more efficient while achieving better performance. Moreover, since the GR-1 and G1 embodiments in~\cref{tab:gr1_g1_evaluation} were not available during \ac{lam} pre-training, the performance gains demonstrate that such alignment can be efficiently achieved during only the post-training stage, further validating the data-efficiency as discussed in~\cref{sec:exp:lara_vla}.

\subsection{LARA for Fast Adaptation and Generalization}
\label{sec:exp:real}
\paragraph{Experimental Setup} To further validate the data-efficiency and generalizability of LARA, we evaluate models by pre-training only on OXE data and post-training on \textit{GR1-Sim-24(30)} and \textit{G1-Real(50)} following the \textit{OXE-Constrained Comparison} setting. As both datasets involve embodiments absent from OXE and limited demonstration data, this experiment tests the adaptability and generalizability of LARA for new embodiments and tasks. We provide additional training details in~\cref{app:details:training}.

\paragraph{Results \& Analyses} 
As shown in the top section of~\cref{tab:gr1_g1_evaluation}, we observe a tremendous performance improvement of LARA when adapting models pre-trained on OXE to novel embodiments and tasks, achieving over $\sim$30\% performance improvements on average. This demonstrates that LARA learns embodiment-agnostic action representations from visual semantics which supports fast adaptation to novel embodiments and tasks rather than overfitting to embodiment-specific patterns, enabling stronger generalization capabilities. Nevertheless, a noticeable gap remains when compared to large-scale pre-trained models that have already seen the GR-1 and G1 embodiments, highlighting the importance of extensive pre-training. Even so, under limited demonstration settings, LARA still outperforms the vanilla GR00T-N1.6 baseline on \textit{G1-Real(50)}, indicating strong potential when combined with larger-scale pre-training. Due to computational constraints, we leave the full-scale pre-training on all available open-source datasets to future work.
\begin{table*}[t!]
    \centering
    
        \caption{\textbf{LAM and LARA-LAM Comparison in SIMPLER Evaluation.}}
    \label{tab:motogpt_tok_compare}
    \begin{tabular}{cc c c  c c c}
    \toprule
    Methods & Pick Object & Move Near & Open Drawer & Close Drawer & Pick Coke Can & Avg. \\
    \midrule
     LAM & 36.3 & 61.0 & 25.7 & 38.0 & 53.0 & 42.8\\
    LARA-LAM & \textbf{41.0} & \textbf{63.7} & \textbf{29.3} & \textbf{53.7}  & \textbf{59.7} & \textbf{49.5}\\
    \textit{LARA-\ac{lam} Improvement} & \textcolor{up}{+12.9\%} & \textcolor{up}{+4.4\%} & \textcolor{up}{+14.0\%} & \textcolor{up}{+41.3\%} & \textcolor{up}{+12.6\%} & \textcolor{up}{+15.7\%} \\
     \bottomrule
    \end{tabular}
\end{table*}

\begin{figure*}
    \begin{minipage}[c]{0.63\linewidth}
    \centering
    \includegraphics[width=0.9\linewidth]{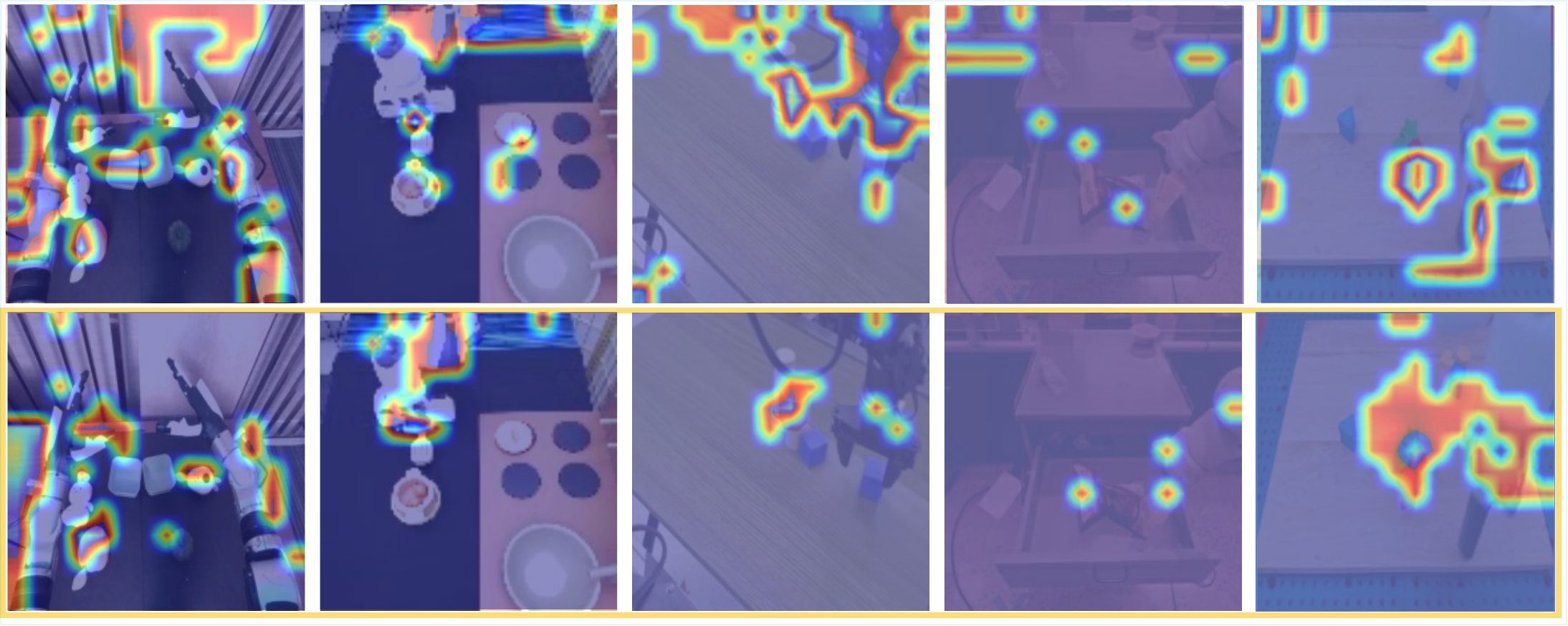}
    \caption{\textbf{Attention Map Visualization for LAM and LARA-LAM.}  We show attention heat maps between latent actions and image patches from \ac{lam} (up) and LARA-LAM (below) respectively, higher attention regions are marked in red.}
    \label{fig:lam_visualization}
    \end{minipage}
    \hfill
    \begin{minipage}[c]{0.35\linewidth}
    \centering
    \includegraphics[width=1.0\linewidth]{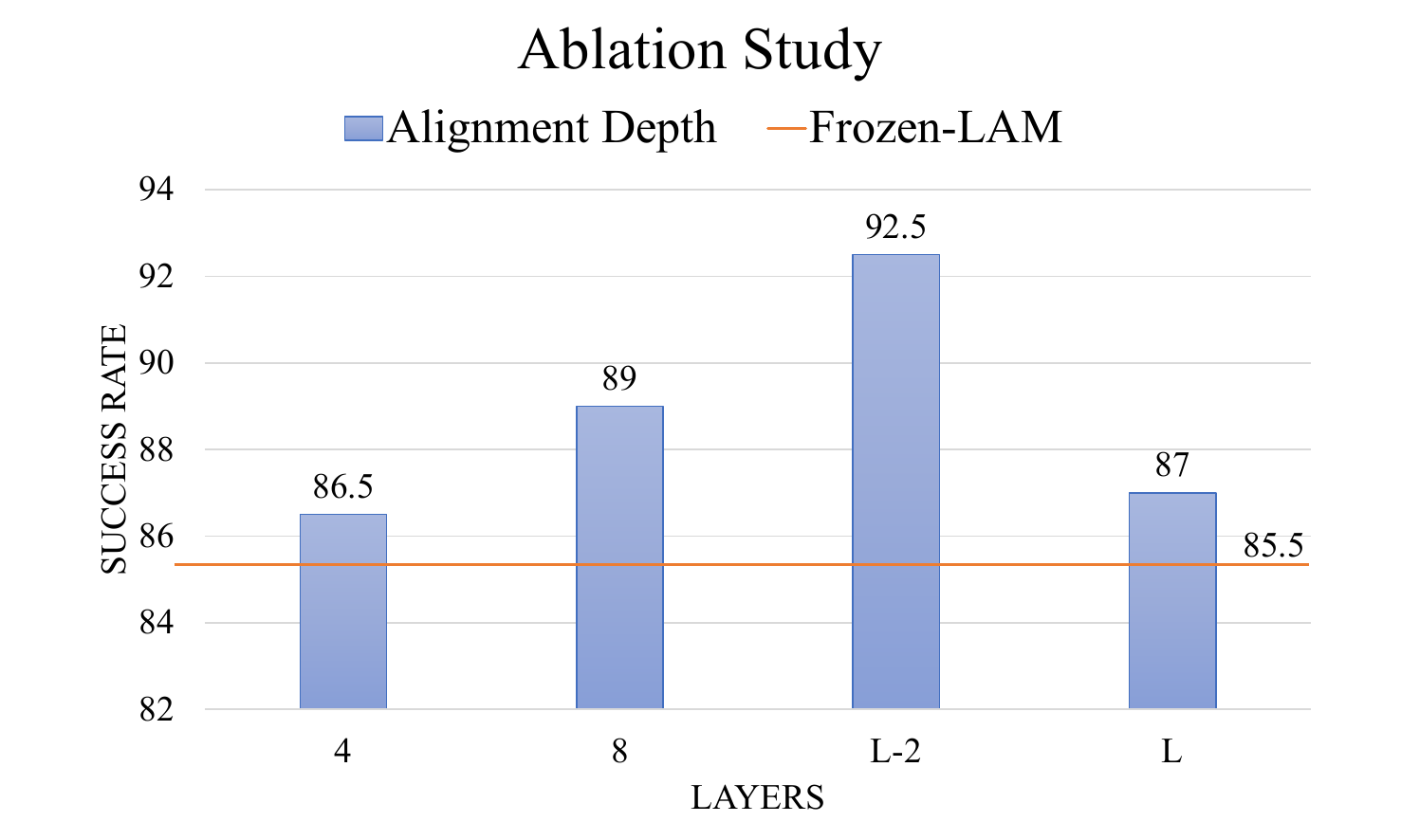}
    \caption{\textbf{Ablation Study on LARA Design.} We report success rates on \textit{LIBERO-Long}, the most challenging subset of LIBERO benchmark.}
    \label{fig:ablation_layer}
    \end{minipage}
    \vspace{-15pt}
\end{figure*}

\subsection{LARA for Latent Action Refinement}\label{sec:exp:lam}
\paragraph{Experimental Setup} To verify the reciprocal enhancement LARA provides to the \ac{lam} model, we investigate whether the alignment process improves the quality of the latent action representations for downstream tasks. We utilize the Moto-GPT~\cite{chen2024moto} framework as a controlled testbed, leveraging its reliance on \ac{lam}-generated latent action tokens for \ac{vla} supervision. Specifically, Moto-GPT employs a two-stage curriculum, where an initial \ac{lam}-only training phase supervise \ac{vla} models exclusively by pseudo-labels from \acp{lam}, followed by a joint training phase that combines latent action supervisions with ground-truth action supervisions. We conduct a direct comparison of \acp{lam} by training two distinct Moto-GPT models on the OXE Fractal dataset from scratch and testing on the \textit{SIMPLER-ENV} benchmark, utilizing latent tokens from a vanilla \ac{lam} and a LARA-aligned \ac{lam} (LARA-LAM) both trained on OXE data, respectively. We provide additional training details in~\cref{app:lam_exp}.

\paragraph{Quantitative \& Qualitative Analyses} As shown in~\cref{tab:motogpt_tok_compare}, LARA-LAM outperforms the baseline LAM across all tasks, yielding an average success rate improvement of 15.7\%. Given that our controlled training protocol isolates representation quality as the sole variable, these performance gains are directly attributable to the superior structure of the LARA-refined latent space. This empirically confirms that the alignment objective is bi-directional: it does not merely transfer knowledge from LAM to Policy, but establishes a reciprocal cycle where action supervision actively refines the \ac{lam} into a more robust, action-centric manifold. To qualitatively validate this refinement, we visualize the attention map between latent action tokens and patch embeddings in both models. As illustrated in~\cref{fig:lam_visualization}, the LARA-LAM attention maps demonstrate a significantly sharper focus on the robot's end-effector and interaction targets, whereas the baseline frequently attends to background distractors. This visual evidence corroborates that LARA functions as an effective inverse dynamics regularizer, suppressing visual noise to prioritize task-relevant motion features.

\subsection{Ablation Study}
\label{sec:exp:ablation}
To validate our design choices, we conduct ablations on the \textit{LIBERO-Long} benchmark using the \textit{Unconstrained} GR00T-N1.6-LARA pipeline following the protocol of \textit{LIBERO-Long} Evaluation in~\cref{sec:exp:lara_post}.

\paragraph{Alignment Depth.} We first investigate the optimal depth for injecting the LARA alignment loss in GR00T-N1.6 model. Given the DiT layers $\textbf{L}$, We evaluate alignment performance at various depths (specifically layers $\{4, 8, $\textbf{L}-2$, $\textbf{L}$\}$). As shown in~\cref{fig:ablation_layer}, the earlier layers lack sufficient semantic abstraction, while the final layers are too specialized. Consequently, we select the $\textbf{L}-2$ layer as the optimal insertion point, offering the balance between high-level semantics and actionable motion. 

Notably, this result does not imply that $\mathbf{L}-2$ is universally optimal across different backbone architectures. To further examine the effect of backbone architecture on alignment depth, we additionally evaluate LARA on the $\pi_{0.5}$ model, with details provided in~\cref{app:pi0.5_exp}. The results show that, for $\pi_{0.5}$, applying alignment at the final layer  achieves the best performance, whereas applying alignment at layer $\mathbf{L}-2$ leads to degraded performance. These findings suggest that the optimal alignment depth is architecture-dependent. Nevertheless, our empirical results indicate a consistent principle: LARA alignment is generally more effective in deeper layers close to the action prediction head, rather than in early layers with limited semantic abstraction.

\paragraph{Joint Optimization vs. Frozen LAM.} 
We further analyze the impact of jointly optimizing LAM alongside the flow-based VLA model, versus using a frozen LAM as a fixed supervision target, both aligning at DiT $\textbf{L}-2$ layer. Results in~\cref{fig:ablation_layer} demonstrate that the joint optimization strategy outperforms the frozen baseline. This empirical evidence supports our core thesis: the bidirectional information flow, where the policy informs the LAM and vice versa, is critical for maximizing performance. Additional analyses of the loss design and ablations on the loss weights $w_1$ and $w_2$ are provided in~\cref{app:ablate_exp}. The ablation results demonstrate that, beyond the joint training strategy, the specific loss design also plays a crucial role.

%% file: Sections/conclusion.tex
\section{Conclusion}
\label{sec:conclusion}
In this work, we introduced \ac{lara} (\textbf{L}atent \textbf{A}ction \textbf{R}epresentation \textbf{A}lignment), a novel framework designed to  co-align the latent action space with the policy's internal representations, overcoming the scarcity of robot action data.
This alignment unlocks a critical reciprocal benefit: the \ac{lam} is grounded by real action trajectories, and the \ac{vla} is regularized by the \ac{lam}'s forward dynamics priors. We demonstrated the versatility of \ac{lara} across multiple paradigms, including pre-training from scratch, post-training enhancement, and latent space refinement based on extensive experimental results.  While our experiments were conducted on subsets of the OXE dataset due to computational constraints, the robust gains observed even in this data-limited regime, highlight the scalability of our approach. We hope this study serves as a foundational guide for future research into the end-to-end co-training of world models and policies, unlocking the full potential of internet-scale video data for generalist robot learning.

%% file: Sections/appendix.tex
\appendix
\onecolumn
\crefname{section}{Appendix}{Appendices}    
\Crefname{section}{Appendix}{Appendices}    
\setcounter{table}{0}
\setcounter{figure}{0}
\setcounter{footnote}{0}
\renewcommand{\thefigure}{S.\arabic{figure}}
\renewcommand{\thetable}{S.\arabic{table}}
\renewcommand{\theequation}{S.\arabic{equation}}
\section{Training Details}
\label{app:method}
\subsection{Model Implementation}
\label{app:details}
\paragraph{Latent Action Model (LAM) Implementation}
We adopt the architectural design from Moto-GPT~\cite{chen2024moto} for our Latent Action Model. The process operates in four stages:
\begin{itemize}[leftmargin=*]
\item \textbf{Visual Encoding:} The current frame $I_t$ and the target future frame $I_{t+C}$ are processed by a frozen, pre-trained ViT~\cite{he2022masked} encoder to extract patch embeddings. These embeddings are concatenated to form a unified visual feature sequence.
\item \textbf{Motion Extraction (M-Former):} These features are input to the "M-Former," a 4-layer transformer encoder equipped with $8$ learnable query embeddings. The M-Former utilizes self-attention to distill the visual changes into a continuous latent representation $z_t$.
\item \textbf{Quantization:} The output query features are discretized using a Vector Quantization (VQ) codebook with a vocabulary size of $128$, resulting in discrete latent motion tokens $z_t^q$.
\item \textbf{Reconstruction:} Finally, the quantized tokens $z_t^q$ are fed into a decoder—comprising a 12-layer ViT~\cite{he2022masked} with a hidden size of $768$—to reconstruct the future frame $I_{t+C}$.
\end{itemize}

\paragraph{Flow-based VLA Implementation}
We employ Eagle-2~\cite{li2025eagle} as the vision-language backbone to process visual observations and task instructions. While the core VLM weights remain frozen, we introduce a trainable self-attention adapter to refine the VLM embeddings before they condition the diffusion process.
To handle diverse robot morphologies, the action policy utilizes embodiment-specific MLP encoders. These encoders project proprioceptive states and noisy actions into a shared latent embedding space. The diffusion process is modeled by a Diffusion Transformer (\ac{dit}) comprising $L=16$ layers, featuring alternating self-attention and cross-attention blocks (conditioned on VLM embeddings).
To support scalability across diverse hardware, we instantiate the model with a capacity for up to $64$ distinct embodiment IDs. We set the maximum action dimension to $32$ and the state dimension to $64$, using padding to accommodate the varying degrees of freedom found in diverse manipulation datasets.

We list the implementation details of each component in ~\cref{tab:implement_details}.
\begin{table}[t!]
    \centering
        \caption{\textbf{Implementation Details of LAM and Diffusion-based VLA Policy with LARA. }}
    \begin{tabular}{l|c c}
    \toprule
      Component   & Parameter & Value   \\
       \rowcolor[HTML]{F0F7FF}\multicolumn{3}{l}{\footnotesize{\textit{LAM}}}\\
      ViT Encoder   &  - & Pretrained ViT\\
      \midrule
      \multirow{2}{*}{M-Former} & num\_queries & 8 \\
      & num\_layers & 4 \\
      \midrule
     \multirow{2}{*}{ViT Decoder} & num\_layers & 12 \\
      & num\_heads & 12 \\
      \midrule
      VQ Codebook & num\_codes & 128 \\
      \midrule 
      \midrule
       \rowcolor[HTML]{F0F7FF}\multicolumn{3}{l}{\footnotesize{\textit{VLA Policy}}}\\
      VLM & - & Frozen Eagle v2\\
      \midrule
       \multirow{2}{*}{Adapter} & Self-Attn & 1 \\
       & Layer Norm & 1 \\
       \midrule
       Action Encoder & MLP & 1\\
       Action Decoder & MLP & 1 \\
       State Encoder & MLP & 1 \\
      Diffusion Model & DiT &  16 \\
      Projector & MLP & 1\\
      Alignment Depth & DiT Layer & L-2\\
      -  & Max Num Embodiments   & 64\\
      - & Max State Dim  & 64 \\
      - & Max Action Dim  & 32 \\
      \bottomrule
    \end{tabular}
    \label{tab:implement_details}
\end{table}

\subsection{Model Training}
\label{app:details:training}
\noindent \textbf{Training Details:} We train the \ac{lam} on subsets of the OXE dataset~\cite{o2024open}, with the temporal stride ($C=5$) to learn the latent action every 5 frames with standard VQ-VAE~\cite{van2017neural} objective. The model is trained for $350$k steps on 4 NVIDIA A100 GPUs with a global batch size of $512$. We utilize the AdamW optimizer with a peak learning rate of $1\times 10^{-4}$ and a cosine decay schedule (weight decay $1\times 10^{-5}$). For further architectural specifics, we refer readers to~\cite{chen2024moto}.

\paragraph{LARA (\textit{DiT-only}) Training}
For the baseline diffusion training, we optimize the flow-matching objective defined in~\cref{eq:action_loss}.  To predict continuous action, we usually predict the action chunk $\textbf{A}_{t:t+C}$. We set $C=16$. We train on the subset of the OXE dataset containing valid action labels. The action prediction horizon is set to $16$. The model is trained for $200$k steps on 4 NVIDIA A100 GPUs with a global batch size of $384$. We use the AdamW optimizer with a peak learning rate of $1\times 10^{-4}$, a cosine decay schedule, and a weight decay of $1\times 10^{-5}$.

\paragraph{LARA (\textit{full}) Joint Training}
For the full LARA framework, we optimize the joint objective in~\cref{eq:full_objective}. The LARA loss is implemented between $z_t$ extracted from ($I_t, I_{t+C}$), with $C=16$, and aligned with $L-2$ layer of hidden states $h_t^\theta$. Crucially, to ensure temporal consistency, we select the hidden state token corresponding to the final timestep of the action chunk ($t+C$) for alignment. This constraint forces the policy's representation of the completed action trajectory ($\textbf{A}_{t:t+C}$) to match the visual effect predicted by the \ac{lam}. Based on empirical tuning, we set the loss balancing weights to $w_1 = 0.01$ and $w_2 = 0.01$. All other optimization hyperparameters (batch size, learning rate, optimizer) remain identical to the \textit{DiT-only} configuration to ensure a fair comparison.

\paragraph{GR00T-N1.6-LARA Post-training.}
In the \textit{Unconstrained} setting, we initialize the model with the public GR00T-N1.6~\cite{bjorck2025gr00t} checkpoint (pre-trained on large-scale data) and perform joint optimization using the protocol described above. We maintain the weights $w_1 = 0.01$ and $w_2 = 0.01$. The model is fine-tuned on target robot demonstrations for approximately $20$k steps, with the exception of the \textit{GR1-Sim-24(30)} benchmark, which is trained for $50$k steps due to higher task complexity. We use a learning rate of $1\times 10^{-4}$ and a global batch size of $384$ on 4 NVIDIA A100 GPUs. This post-training setup is consistent across LARA (\textit{full}), LARA (\textit{DiT-only}), and the GR00T-N1.6 baseline.

We summarize the training hyperparameters in ~\cref{tab:hyperparameters}.

\subsection{Training Dataset}
\label{app:data_details}
\begin{figure}[t]
    \centering
    \begin{minipage}[c]{0.55\linewidth}
        \centering
        \includegraphics[width=0.9\linewidth]{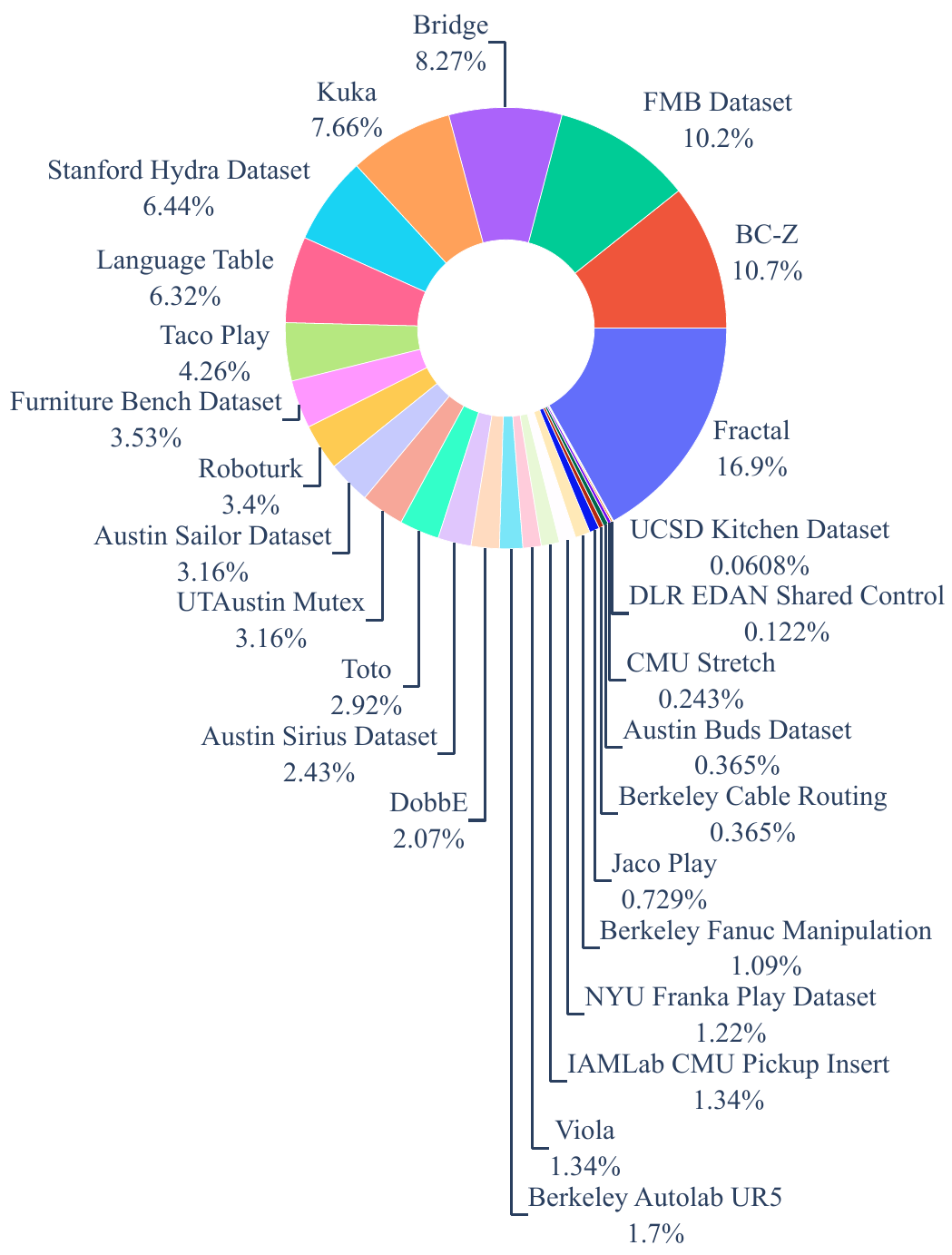}
        \caption{\textbf{Training Data Distribution.} Visualization of the dataset mixtures used for both \ac{lam} pre-training and LARA policy training.}
        \label{fig:train_data_vis}
    \end{minipage}
    \hfill
    \begin{minipage}[c]{0.4\linewidth}
        \centering
         \captionof{table}{\textbf{Training Hyperparameters for models.}}
         \resizebox{0.75\linewidth}{!}{
    \begin{tabular}{c|c}
    \toprule
     Parameter &  Value \\
     \midrule
      \rowcolor[HTML]{F0F7FF}\multicolumn{2}{l}{\footnotesize{\textit{LAM Pre-train}}}\\
    Batch Size & 512 \\
      Optimizer & AdmaW \\
      LR\_max & 1e-4 \\
      LR\_schedule & cosine decay \\
      Weight\_decay & 1e-5 \\
      Training Steps & 350K \\
      NUM of GPUs & 4 A100\\
      \midrule
       \rowcolor[HTML]{F0F7FF}\multicolumn{2}{l}{\footnotesize{LARA (~\textit{full} / ~\textit{DiT-only})}}\\
           Batch Size & 384 \\
      Optimizer & AdamW \\
      LR\_max & 1e-4 \\
      LR\_schedule & cosine decay \\
      Weight\_decay & 1e-5 \\
      Training Steps & 200K \\
      NUM of GPUs & 4 A100\\
      \midrule
         \rowcolor[HTML]{F0F7FF}\multicolumn{2}{l}{\footnotesize{\textit{Post-training}}}\\
           Batch Size & 384 \\
      Optimizer & AdamW \\
      LR\_max & 1e-4 \\
      LR\_schedule & cosine decay \\
      Weight\_decay & 1e-5 \\
      Training Steps & 20K / 50K \\
      NUM of GPUs & 4 A100\\
      \bottomrule
    \end{tabular}
    }
    \label{tab:hyperparameters}
    \end{minipage}
\end{figure}

We curate a targeted subset of the Open XEmbodiment (OXE) dataset~\cite{o2024open}, specifically filtering for trajectories featuring single-arm end-effector control. The detailed composition and distribution of these subsets are visualized in~\cref{fig:train_data_vis}.
To tailor the data for distinct learning objectives, we employ different temporal strides ($C$). For \ac{lam} pre-training, we set a shorter horizon of $C=5$ to capture fine-grained visual motion dynamics. Conversely, for the VLA policy training, we extend the horizon to $C=16$ to match the length of the predicted action chunks. To address the variance in dataset sizes within OXE, we adopt a balanced sampling strategy where each subset is sampled with equal probability, preventing the model from overfitting to dominant data sources.

\section{Additional Experimental Results}
\label{app:additional_results}
\subsection{$\pi_{0.5}$ Post-training with LARA}
\label{app:pi0.5_exp}
To further validate the effectiveness of LARA as a plug-and-play module, we integrate LARA with the pretrained $\pi_{0.5}$~\cite{black2025pi_} model. We post-train $\pi_{0.5}$-LARA on the LIBERO dataset for 20k steps, following the post-training recipe of $\pi_{0.5}$. The LARA alignment loss is applied to the final layer, \eg, layer $\mathbf{L}$, of the $\pi_{0.5}$ backbone, immediately before the action decoder. We use loss weights $w_1 = 0.01$ and $w_2 = 0.01$. Additionally, we evaluate applying the alignment loss at layer $\mathbf{L}-2$, with results reported in~\cref{tab:pi0.5_result}.

As shown in~\cref{tab:pi0.5_result}, LARA consistently improves upon the already strong performance of the base model across the LIBERO benchmark. Moreover, the optimal alignment depth depends on the backbone architecture and must be chosen carefully. While the exact optimal layer index varies across architectures, our empirical findings suggest a consistent principle: alignment is more effective in deeper layers close to the action prediction head.

\begin{table*}[t!]
    \centering
        \caption{\textbf{Comparison of $\pi_{0.5}$ and $\pi_{0.5}$-LARA in LIBERO.}}
    \label{tab:pi0.5_result}
    \begin{tabular}{l | c c c  c | c }
    \toprule
    Methods & Spatial &	Object &	Goal &	Long &	Average \\
    \midrule
    $\pi_{0.5}$ & 98.8	& 98.0 & 98.2 &	92.4 &	96.9 \\
    $\pi_{0.5}$-LARA (L-2 layer)	&97.0 &	99.0	&87.5 &	83.5	& 91.2\\
$\pi_{0.5}$-LARA (L layer)	& \textbf{99.0} &	\textbf{98.5} &	\textbf{99.0} &	\textbf{94.5} &	\textbf{97.8} \\
\textit{\ac{lam} Improvement} &  \textcolor{up}{+0.2\%} & \textcolor{up}{+0.5\%} & \textcolor{up}{+0.8\%} & \textcolor{up}{+2.1\%} &  \textcolor{up}{+0.9\%} \\
     \bottomrule
    \end{tabular}
\end{table*}

\subsection{Loss Designs and Weights Ablation}
\label{app:ablate_exp}

We further ablate the use of the LARA loss and the LAM loss to verify the importance of the loss design. The experimental setup follows the ablation study in~\cref{sec:exp:ablation}. Specifically, we evaluate GR00T-N1.6-LARA on the LIBERO-Long dataset. In addition, we conduct an ablation study on the loss weights, varying them from 0.0001 to 1.0. The results are presented in~\cref{app:loss_ablate}.

As shown in~\cref{app:loss_ablate}, joint training with both the LARA loss and the LAM loss achieves the best performance. The LARA loss is essential, as it regularizes the feature space and enables the policy and the LAM to mutually enhance each other, thereby achieving the strongest performance. For the loss-weight ablation, the optimal setting is $w_1 = 0.01$ and $w_2 = 0.02$. In contrast, larger weights, such as 0.1 or 1.0, degrade action prediction accuracy, as the LARA alignment loss and the LAM loss begin to dominate model training. Unless otherwise specified, we use weights of 0.01 across all tasks and settings.

    

\section{Details on Real World Experiments}
\label{app:g1_exp}
\begin{figure}
\begin{minipage}[c]{0.4\linewidth}
\centering
            \caption{\textbf{Loss Designs and Weights Ablation.} We evaluate various weights and the two regularization losses in the LIBERO-Long dataset.}
    \label{app:loss_ablate}
    \begin{tabular}{l |c }
    \toprule
    Methods & Long \\
    \midrule
    $w_1=0.0001$ \& $w_2=0.0001$ & 91.5\\
    $w_1=0.001$ \& $w_2 = 0.001$ & 91.0 \\
    $w_1=0.01$ \& $w_2=0.01$ & \textbf{92.5} \\
    $w_1=0.1$ \& $w_2=0.1$ & 89.5 \\
    $w_1=1.0$ \& $w_2=1.0$ & 86.5 \\
    \midrule
    w/o $\mathcal{L}_{LARA}$ & 88.0 \\
    w/o $\mathcal{L}_{LAM}$ & 89.5 \\
    \bottomrule
    \end{tabular}
\end{minipage}
\hfill
\begin{minipage}[c]{0.6\linewidth}
        \centering
    \includegraphics[width=0.6\linewidth]{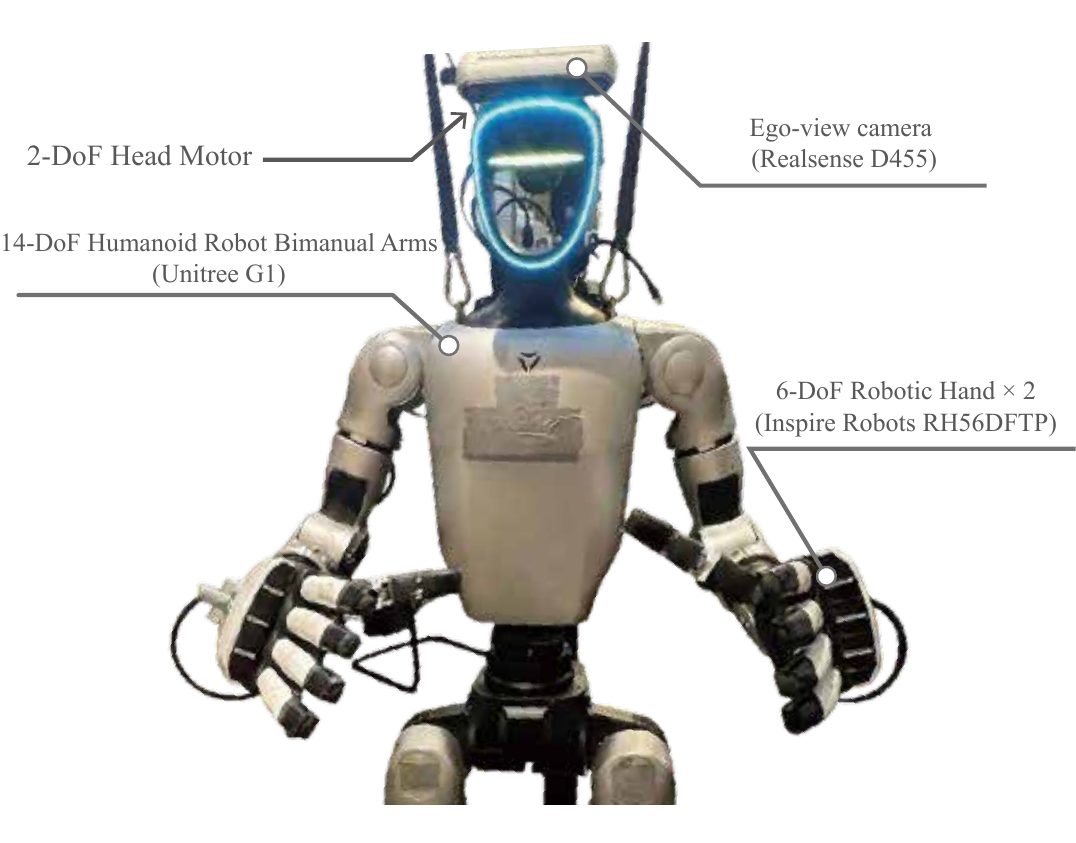}
    \caption{\textbf{Real-world setup with the Unitree G1 humanoid.} The robot is equipped with Inspire Hands and a 2-DoF actuated head mounting an Intel RealSense D455 RGB camera for first-person observations.}
    \label{fig:g1_setup}
\end{minipage}

\end{figure}

\paragraph{Hardware.}
All real-world experiments are conducted on a Unitree G1 humanoid equipped with Inspire Hands and a actuated head for camera reorientation (shown in \cref{fig:g1_setup}). To increase the first-person field of view (FoV), we use the head-mounted Intel RealSense D455 to capture RGB observations.

\paragraph{Control Interface.}
All policy operates on a 28-dimensional state and outputs a 28-dimensional action. The controlled DoFs consist of 14 upper-body arm DoFs (7 DoFs per arm), 12 hand DoFs (6 DoFs per hand), and 2 head DoFs (yaw and pitch). During experiments, the robot is suspended by a gantry crane for safety and stability.

\begin{figure*}
    \centering
    \includegraphics[width=1.0\linewidth]{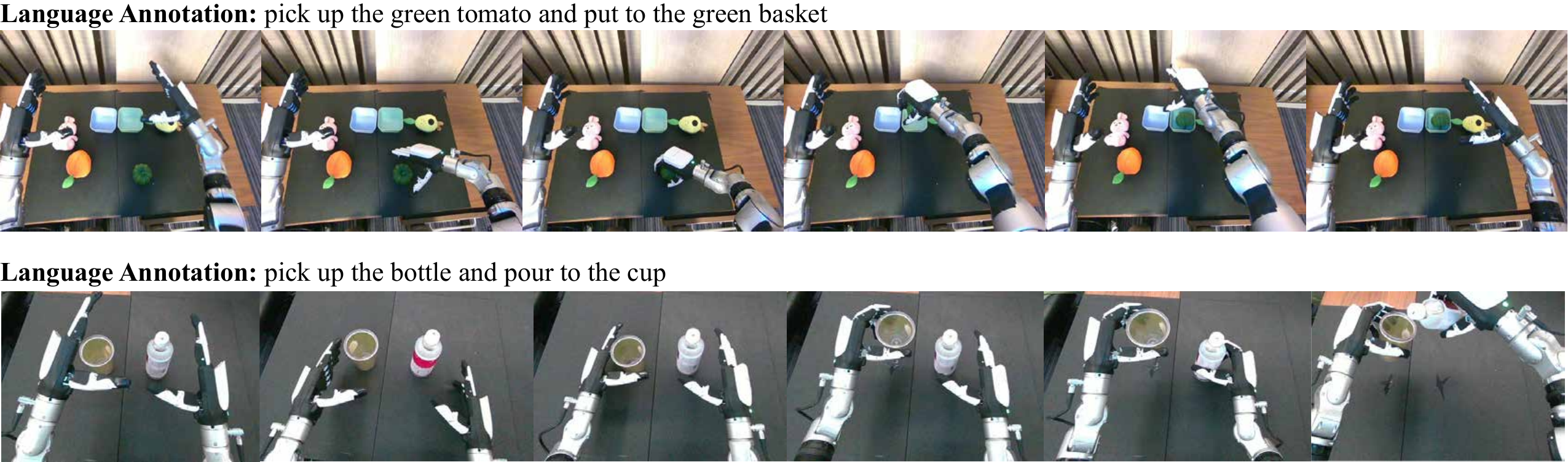}
    \caption{Visualization of real-world tasks.}
    \label{fig:g1_data}
    \vspace{-1em}
\end{figure*}

\paragraph{Data collection.}
We collect real-world VLA training data by teleoperating the G1 using Apple Vision Pro. We consider two manipulation tasks: (i) pick-and-place and (ii) pouring. For each task, we collect 50 demonstrations (examples shown in Fig.~\ref{fig:g1_data}).  For both tasks, the object poses are randomized within a $10\text{ cm} \times 10\text{ cm}$ region.

\paragraph{Tasks Evaluation Metrics.} For the two tasks, each task is evaluated over 50 trials. The inference denoising step is 4, and the action horizon is 8.

\textbf{(1) Single-Arm Pick-and-Place.}
\textit{Instruction: "Pick the Green Tomato and Place in the Green Basket."}
This task evaluates the sim-to-real transfer capability of LARA (\textit{full}) when trained only on OXE data, compared to the GR00T-N1.6 baseline which benefits from large-scale pre-training (see in ~\cref{tab:gr1_g1_evaluation}). Success is decomposed into stages:
\begin{itemize}[leftmargin=*,noitemsep]
\item \textbf{Pick ($\text{SR}_{\text{Pick}}$):} The robot successfully grasps and lifts the tomato (e.g., Frame 4 in~\cref{fig:g1_data}).
\item \textbf{Full ($\text{SR}_{\text{Full}}$):} The tomato is successfully placed into the target basket (e.g., Frame 5).
\item \textbf{Place ($\text{SR}_{\text{Place}}$):} We define the conditional success rate for the placement phase ($\text{SR}_{\text{Place}}$) as the probability of success given a successful pick:
\begin{equation}
\text{SR}_{\text{Place}} = \frac{\text{SR}_{\text{Full}} - \text{SR}_{\text{Pick}}}{\text{SR}_{\text{Pick}}}.
\end{equation}
\end{itemize}

\textbf{(2) Bimanual Pouring.} \textit{Instruction: "Grasp the Bottle and Pour to the Cup."} This task serves as a challenging benchmark for bimanual coordination, with varying embodiment gap, task gap compared with our training dataset. Success is tracked for each effector:
\begin{itemize}[leftmargin=*,noitemsep]\item \textbf{Grasp-Left ($\text{SR}_{\text{GL}}$):} The left hand successfully grasps the cup (e.g., Frame 4).
\item \textbf{Grasp-Right ($\text{SR}_{\text{GR}}$):} The right hand successfully grasps the bottle (\eg, Frame 5).
\item \textbf{Full ($\text{SR}_{\text{Full}}$):} Liquid (or proxy object) is successfully poured from the bottle to the cup (e.g., Frame 6).
\item \textbf{Pour ($\text{SR}_{\text{Pour}}$):} Since the pouring action requires the successful execution of both grasps, we define the conditional success rate for pouring ($\text{SR}_{\text{Pour}}$) as:
\begin{equation}
\text{SR}_{\text{Pour}} = \frac{\text{SR}_{\text{Full}} - \text{SR}_{\text{GL}} \times \text{SR}_{\text{GR}} }{\text{SR}_{\text{GL}} \times \text{SR}_{\text{GR}}}.
\end{equation}
\textit{Note: We assume independence between the grasp success probabilities for the normalization factor.}

\end{itemize}

\section{LARA-LAM for Latent Action Refinement}
\label{app:lam_exp}

\paragraph{Moto-GPT Vanilla Pipeline.} The vanilla SIMPLER pipeline in Moto-GPT \citep{chen2024moto} consists of three stages: (1) latent tokenizer pre-training, which utilizes a subset of Open-X-Embodiment \citep{o2024open} including 109k real-
world trajectory video covering various embodiments; (2) Moto-GPT pre-training, which utilizes the same Open-X-Embodiment subset to supervise the \ac{vla} model \textbf{only} with latent action tokens; and (3) Moto-GPT co-fine-tuning, which additionally uses 73k expert trajectories with action labels from RT-1 \citep{brohan2022rt}, the loss in Stage-3 contains both the latent action tokens prediction loss and the real action prediction loss. The evaluation on SIMPLER includes three tasks based on the Google-Robot embodiment: Pick Coke Can, Move Near, and Open/Close Drawer.

\paragraph{Our implementation.} We directly begin with our pre-trained \ac{lam} while skipping the first stage of Moto-GPT. We then proceed to perform Moto-GPT stage-2 and stage-3 training with our pre-trained \ac{lam}. Notably, for training efficiency, in this experiment we use a smaller dataset (OXE Fractal dataset) for Moto-GPT stage-2 pre-training, and stage-3 co-fine-tuning. We compare the following two variants:
\begin{itemize}[leftmargin=*,nolistsep]
    \item \textbf{LAM.} We directly use our stage-1 pre-trained \ac{lam} to guide the Moto-GPT stage-2 pre-training, and then continue to perform Moto-GPT stage-3 co-fine-tuning in OXE Fractal dataset.
    \item \textbf{LARA-LAM.} We use our trained LARA-LAM for the training of Moto-GPT stage-2 and stage-3 in OXE Fractal dataset. In contrast to the vanilla LAM, the LARA-LAM undergoes an extra LARA \textit{joint pre-training} stage to verify whether LARA joint pre-training yields better latent action representation.
\end{itemize}
